\documentclass{trbunofficial}
\usepackage{latexsym}
\usepackage{graphicx}
\usepackage{amsfonts,amssymb,amsmath}
\usepackage{hyperref}

\usepackage{graphics} 
\usepackage{dblfloatfix} 
\usepackage[ruled,vlined]{algorithm2e}
\usepackage[utf8]{inputenc} 
\usepackage[T1]{fontenc}    
\usepackage{hyperref}       
\usepackage{url}          
\usepackage{booktabs}      
\usepackage{nicefrac}       
\usepackage{microtype}     
\usepackage{caption}
\usepackage{subcaption}
\usepackage{catchfile}
\usepackage{totcount}
\usepackage{multirow}
\usepackage{array}

\newcolumntype{P}{>{\centering\arraybackslash}m{1.45cm}}
\newcolumntype{Q}{>{\centering\arraybackslash}m{0.90cm}}

\title{Computer Vision for Transit Travel Time Prediction: An End-to-End Framework Using Roadside Urban Imagery}

\author{\textbf{Awad Abdelhalim}, Corresponding Author \\
  Postdoctoral Research Associate \\
  Department of Urban Studies and Planning \\
  Massachusetts Institute of Technology, Cambridge, MA 02139, USA \\
  Email: awadt@mit.edu \\
  \hfill\break
  \textbf{Jinhua Zhao} \\
  Associate Professor \\
  Department of Urban Studies and Planning \\
  Massachusetts Institute of Technology, Cambridge, MA 02139, USA \\
  Email: jinhua@mit.edu
}

\nolinenumbers

\begin{document}
\maketitle

\section{Abstract}
Accurate travel time estimation is paramount for providing transit users with reliable schedules and dependable real-time information. This paper is the first to utilize roadside urban imagery for direct transit travel time prediction. We propose and evaluate an end-to-end framework integrating traditional transit data sources with a roadside camera for automated roadside image data acquisition, labeling, and model training to predict transit travel times across a segment of interest. First, we show how the General Transit Feed Specification (GTFS) real-time data can be utilized as an efficient activation mechanism for a roadside camera unit monitoring a segment of interest. Second, Automated Vehicle Location (AVL) data is utilized to generate ground truth labels for the acquired images based on the observed transit travel time percentiles across the camera-monitored segment during the time of image acquisition. Finally, the generated labeled image dataset is used to train and thoroughly evaluate a Vision Transformer (ViT) model to predict a discrete transit travel time range (band). The results of this exploratory study illustrate that the ViT model is able to learn image features and contents that best help it deduce the expected travel time range with an average validation accuracy ranging between 80\%-85\%. We assess the interpretability of the ViT model's predictions and showcase how this discrete travel time band prediction can subsequently improve continuous transit travel time estimation. The workflow and results presented in this study provide an end-to-end, scalable, automated, and highly efficient approach for integrating traditional transit data sources and roadside imagery to improve the estimation of transit travel duration. This work also demonstrates the value of incorporating real-time information from computer-vision sources, which are becoming increasingly accessible and can have major implications for improving operations and passenger real-time information.

\hfill\break%
\noindent\textit{Keywords}: Travel time prediction, transit performance, vision transformers.
\newpage

\section{Introduction}
Accurate and reliable travel time prediction plays a critical role in all aspects of transportation planning. This is even more crucial when it comes to public transit. Efficient operations, riders' experience, and the general perception of public transit are greatly shaped by the on-time performance of the system. Travel and arrival time prediction, in general, and particularly for public transit applications are fields of research that have been well studied over the years. The recent advancements in machine learning and artificial intelligence technologies have significantly improved the state of practice. This is in part due to the wealth of data available for and from day-to-day public transit operations, including high granularity data for passenger count and movements (APC), automated fare collection (AFC), automated vehicle locations (AVL), and the information from the General Transit Feed Specification (GTFS). While these data sources offer an abundance of scheduling, usage, and performance measures that can be used for tasks like travel time prediction for transit, transit vehicles still - for the most part- share road infrastructure with other roadway users. This necessitates combining transit-specific data with more generalized data sources that can offer information about the overall traffic state and roadway infrastructure, which are often challenging to acquire.

The recent advancements in the fields of deep learning, machine perception, and computer vision have made image data extremely useful. Tasks including image classification, detection, and tracking of objects within images can be accomplished with ease,  with an ever-growing multitude of frameworks and architectures to choose from. The biggest drawback when opting to utilize computer vision architectures is that they remain extremely data-hungry, requiring copious amounts of data that needs to be acquired and, more often than not, manually labeled to train vision models. And while this is a challenge facing the broader computer vision community, it is further exacerbated in domain-specific applications like transit operations, where the need to acquire external technology and talent for these tasks of data acquisition and annotation is often cost-prohibitive. The authors of this paper, see immense value in incorporating the domain knowledge of relevant data sources to create a streamlined framework for image data acquisition, labeling, and vision model training combining roadside imagery with transit data sources. This study presents our exploratory framework for this concept and experimental results from a pilot study conducted in a segment of Massachusetts Avenue in Cambridge, MA, USA. 

\subsection{Objective and Contributions}
In this study, we propose and evaluate TranViT, an end-to-end framework for efficiently integrating real-time GTFS, AVL, roadside camera data, and a Vision Transformer (ViT) architecture for predicting transit travel time through a camera-monitored segment. The main contribution of this work is presenting a blueprint for the utilization of traditional transit data sources to acquire, label, and train data for computer vision tasks. We demonstrate the resulting highly accurate and interpretable predictions and discuss their implications for the state of the practice. The following sections of this paper are as follows: (a) a survey of related literature in transit travel time prediction and applications of deep learning and computer vision, (b) a detailed breakdown of the proposed TranViT framework, (c) results and analyses of the case study, and, (d) discussion and conclusions.

\subsection{Related Work}
\subsubsection{General Traffic and Transit Travel Time Prediction}
There is a rich body of literature on travel time prediction, a topic that has been well-studied for years. In the context of general traffic, this often falls into larger-scale traffic state estimation, which along with travel time and speed includes predicting traffic flow and density \cite{wang2005real, yang2005travel, work2008ensemble, yildirimoglu2013experienced}. Those works among others have mostly relied on probe vehicle GPS and loop detector data, which are often limited in size and temporal coverage, noisy, and required the use of complementary techniques to overcome the data quality issues, including particle and Kalman filtering, and machine learning  \cite{wang2008real, chen2011real, jenelius2013travel} in combination with underlying traffic state models and, in some cases, simulation modeling.

For transit, accurate travel time prediction is paramount for providing transit users with reliable schedules and dependable real-time information about their transit vehicles' arrival times, and for the efficient implementation of operation strategies such as transit signal priority \cite{zeng2014real, abdelhalim2018impact}. The methods used in general traffic state estimation and travel prediction, particularly from probe vehicle analyses, may not always translate directly into the context of transit. Albeit sharing the same roadway infrastructure; the vehicle dynamics and operations of transit (continuous start-stopping), among other factors like passenger interactions with operators, influence the movement of transit within traffic streams in a way that is not necessarily reflective of the overall traffic. The need to monitor this level of complexity in size (fleets as opposed to individual vehicles) and operation, however, has resulted in an abundance of transit data sources being available to practitioners. Of the aforementioned data sources, AVL data has particularly been at the center of numerous research efforts. An early study by Cathey et al. \cite{cathey2003prescription} offered a generalized framework to utilize AVL data for making transit arrival time predictions. The growing adoption of AVL systems by transit agencies allowed researchers to develop real-time applications \cite{jeong2005prediction, shalaby2004prediction}. While these studies have demonstrated the ability to obtain accurate travel time predictions from AVL data, the lack of information regarding other influencing factors like overall congestion state and the vehicles' own dwell time were limiting factors. Works by Yu et al. \cite{yu2010hybrid, yu2011bus} have demonstrated that incorporating additional external data sources (e.g. weather data) and using stop-level travel time that captures this variation in dwell time more than the route-level analysis results in improved predictions.

The past decade has witnessed substantial strides in data availability (higher frequency AVL, GTFS-RT, APCs, etc) and real-time assessment and predictive methodologies \cite{park2020assessing, elliott2020modelling, aemmer2022measurement, samal2017speedpro}. While state estimation methods based on Kalman filters, and machine learning models based on boosting and ensemble remained popular \cite{zhang2015gradient, gaikwad2019performance}, deep learning architectures based on recurrent neural networks (RNNs) have also proven high competency in the travel time prediction task. Studies by Zhou et al. \cite{zhou2019learning} and Pang et al. \cite{pang2018learning} concluded that RNN-based models significantly outperform other state-of-the-art methods. The proposed Long Short-Term Memory (LSTM) RNN by Pang et al. achieved a minimum of 10\% improvement in the mean absolute error compared to other traditional models. Han et al. \cite{han2020bus} proposed a method based on position calibration and an LSTM model that accurately predicts transit arrival time, with an error below two minutes for the 8\textsuperscript{th} downstream stop. 


\subsubsection{Computer Vision, Vision Transformers, and Related Works in Transportation Applications}
Computer vision (CV) is a field of artificial intelligence focused on the derivation of useful information from images and image-based data (e.g. videos). This includes tasks like image classification, object recognition, and multi-object recognition and tracking. Albeit being proposed since the early 1990s \cite{lecun1998gradient}, the rapid evolution of Convolutional Neural Networks (CNNs) in the past decade helped establish their place as the undisputed backbone for innumerable architectures that conduct the various aspects of computer vision tasks. These CV-based methods have been widely adopted for a variety of applications in the transportation field, including traffic speed, turn count, and density estimation \cite{buch2011review, abdelhalim2021framework, gokasar2021real}, safety \cite{tageldin2014safety, sayed2013automated, abdelhalim2021real}, and autonomous driving \cite{janai2020computer}. The use of CV in transit applications remains extremely rare, with few applications in railway intrusion detection \cite{wang2021fast}, and a recent study by Sipetas et al. \cite{sipetas2020estimation} who utilized a CV-based video processing component as a part of a larger framework to estimate left-behind subway passengers.

The Vision Transformer (ViT) architecture introduced by Dosovitskiy et al. \cite{dosovitskiy2020image} was inspired by the immense success of transformer architectures in the field of natural language processing (NLP) \cite{vaswani2017attention}. Converse to CNNs, which have been the de-facto method for most computer vision tasks, the vision transformer architecture doesn't introduce the implicit bias of the convolutional and pooling layers present in CNNs, allowing the trained model to better extract global information from images. Although this comes at the cost of requiring more training data, the ViT-based models have been shown to outperform their CNN counterparts at their introduction in 2020. While there has been a tug of war between ViT and CNN-based models since then, ViT-based models have been the standout performer in tasks that are more involved than simple everyday image classification, including but not limited to holography \cite{cuenat2021convolutional}, classification of COVID-19 CT scan images \cite{gao2021covid}, and identifying distracted driving \cite{li2022distracted}. Those studies provide an indication that the global-attention aspect of the ViT architecture allows it to make more accurate predictions using information that could be lost within convolution and pooling layers of a CNN. While CNNs are still useful for transportation-related classification tasks where information is locally-concentrated in an image such as traffic sign classification \cite{zheng2022evaluation}, studies that evaluated vision models for transportation applications where the information needed to make accurate predictions are expected to be sparse across the image show the considerable benefits of using ViT. This includes the work of Liang et al. \cite{liang2022stargazer} demonstrating the ability of ViT in detecting driver distraction, and a study by Abdelraouf et al. \cite{abdelraouf2022using} who demonstrated the ability of a ViT-based architecture to detect rain and roadway surface conditions with an impressive F-1 score of up to 98\%.

Computer vision methods and models have proven to add tremendous value to different fields of science and practice. There remains, however, a significant gap between the abundance of available transit data sources and the integration of these data sources with methods and frameworks to extract complementary information through computer vision. Such integration can provide a significant boost to the state of the practice. The authors believe that this gap is, by and large, due to the demanding process of computer vision models' data acquisition, labeling, and training. There is an immense need for developing generalizable and transferable frameworks that can seamlessly integrate these existing transit and urban data sources, and streamline the data acquisition, model training, and inference processes. We propose TranViT as a pioneering example for this much-needed systems integration task.

\section{Methodology}
\subsection{Site of Study}
\begin{figure}[!ht]
\centering
\includegraphics[width= .50\textwidth]{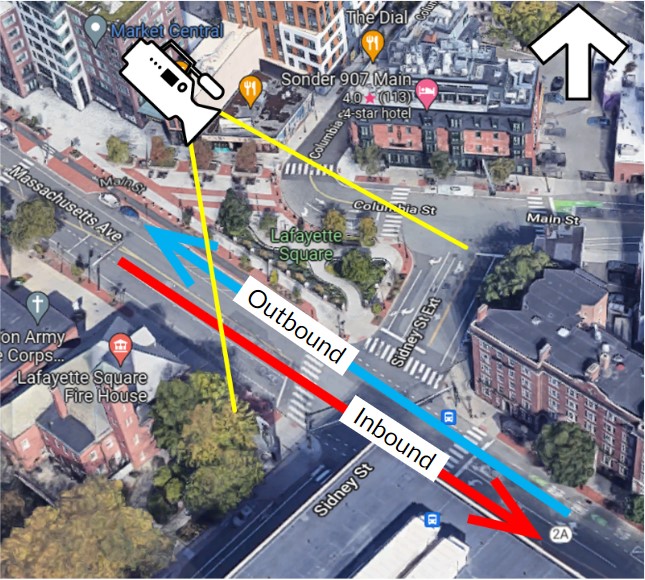}
\caption{\label{fig:site}  Site of study near Central Square in Cambridge, MA, USA.}
\end{figure}

\begin{figure}[!ht]
\begin{subfigure}{.49\textwidth}
  \centering
  \includegraphics[width= \textwidth, height = 2.4in]{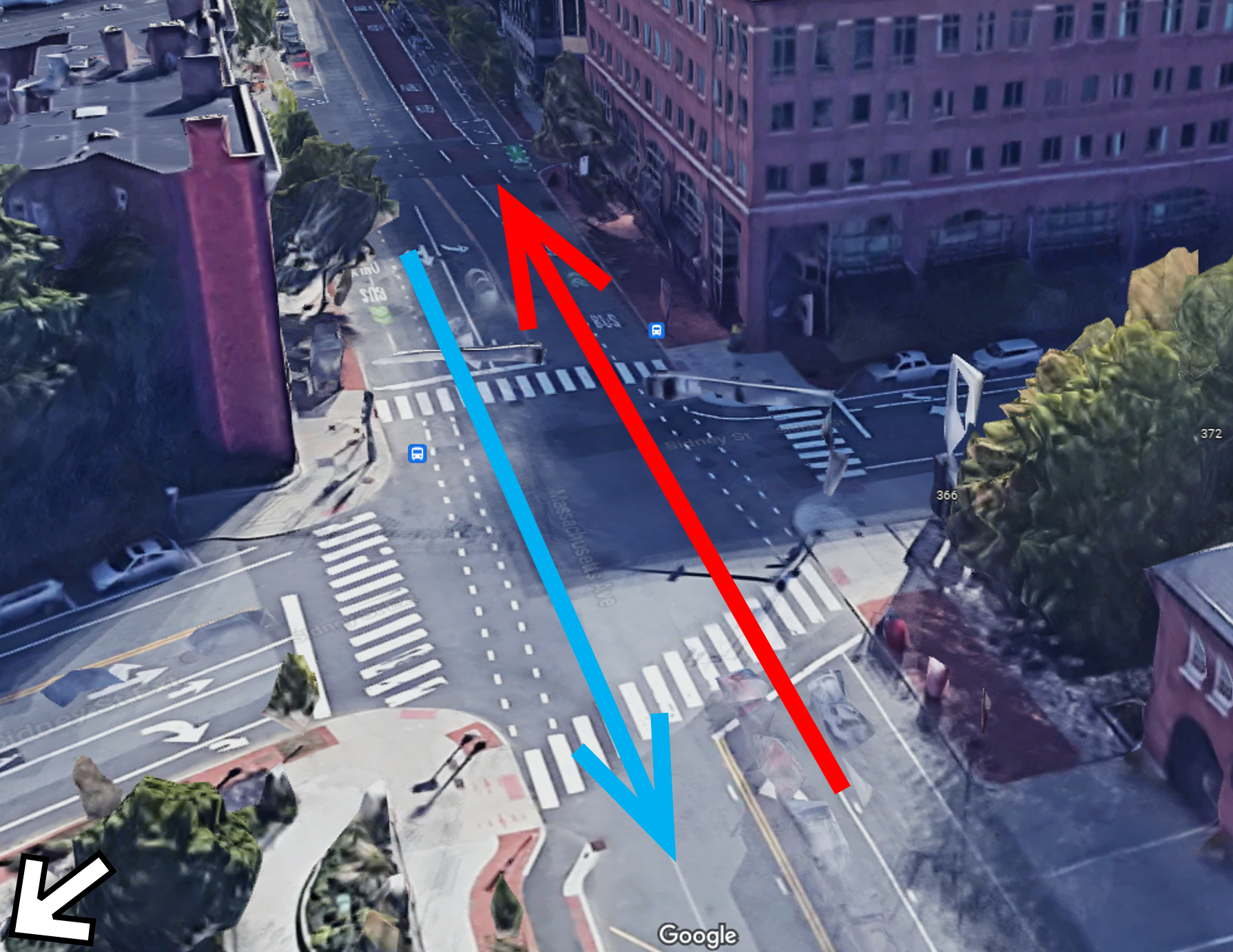}
  \caption{\label{fig:maps_view} Google Maps view.}
\end{subfigure}
\begin{subfigure}{.49\textwidth}
  \centering
  \includegraphics[width= \textwidth, height = 2.4in]{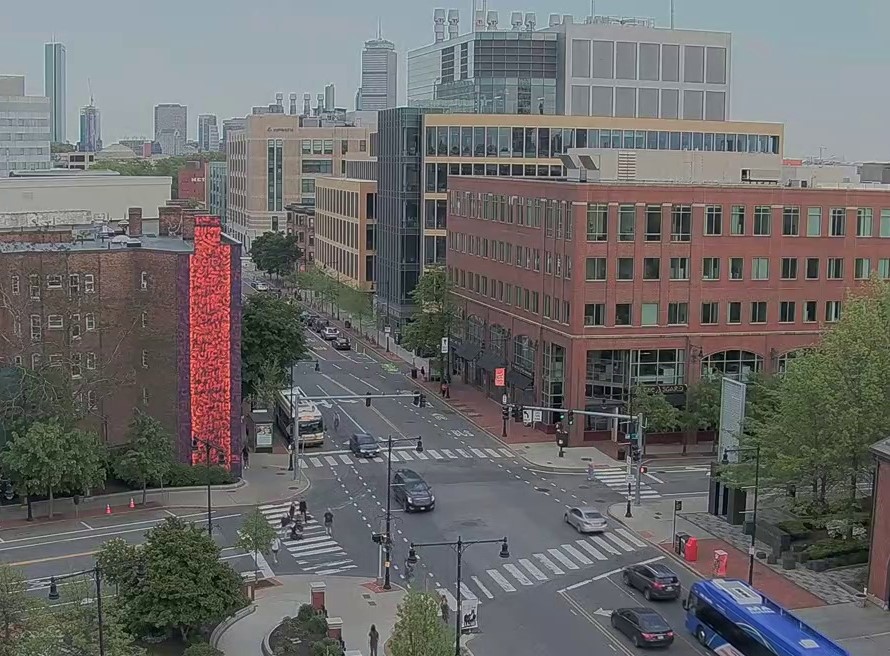}
  \caption{\label{fig:camera_view} Camera view.}
\end{subfigure}
\caption{\label{fig:site_view} Camera perspective for the site of study.}
\end{figure}

The site of this exploratory study was near Central Square in Cambridge, MA, USA, at the intersection of Massachusetts Avenue and Sidney Street. We selected this site due to the availability of a public live camera streaming a live view of this area, which is served by multiple Massachusetts Bay Transportation Authority (MBTA) bus routes. The Google Earth view of the site is illustrated in Figure \ref{fig:site}. The north direction is indicated by the arrow at the top-right of the figure. The top-left shows the position of the public camera that was used to acquire images for this study, with the yellow lines showing the field of view of the camera. The bus stops at the bottom-right quarter of the image serve either direction of the MBTA Route 1 bus operating between Nubian and Harvard Square. The inbound (to Boston) and outbound (to Harvard Sq) directions of the route are respectively illustrated by the red and blue arrows. Figure \ref{fig:site_view} shows the Google Maps view from the camera angle, alongside the true camera perspective of the site.

\subsection{Proposed Framework, Data Collection and Pre-processing, and ViT Model Training}

\begin{figure}[!ht]
\centering
\includegraphics[width= 0.95\textwidth]{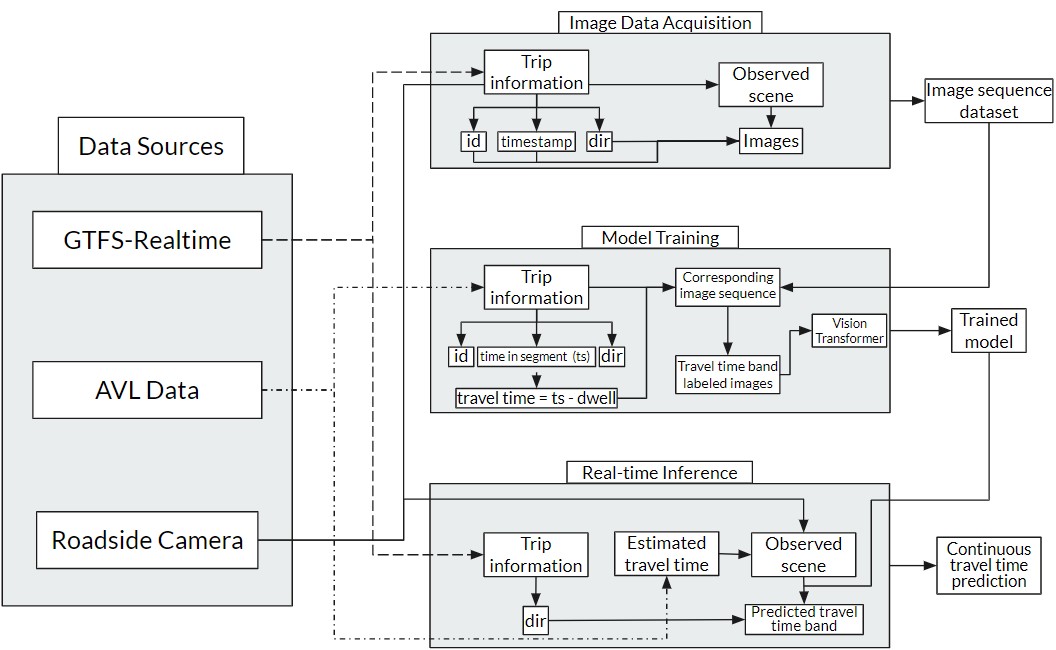}
\caption{\label{fig:framework} TranViT data sources, modules, and workflow.}
\end{figure}

\subsubsection{Data Sources and Training Data Acquisition}
Figure \ref{fig:framework} illustrates the data sources and the modules of our proposed framework. The attributes of data acquired from each source are described in Table \ref{tab:sources}. At the core of our proposed TranViT framework is the General Transit Feed Specification real-time component (GTFS-RT). The GTFS-RT is available for the MBTA vehicles through onboard GPS equipment which updates and publishes vehicle location, heading, and occupancy data in real-time at high-frequency intervals (up to every 1 second) \cite{mbta_gtfs}. We utilize this high-frequency of the GTFS-RT to trigger image acquisition \emph{only} as transit vehicles approach the area of the study, which accomplishes the following:
\begin{enumerate}
    \item Enables linking the acquired image sequences and the travel times associated with the trip ID that activated the image acquisition.
    \item Optimizes the image acquisition process, ensuring data is only collected as needed.
\end{enumerate}

\noindent Image acquisition from camera livestream at the site is activated once a transit vehicle is approaching (within 500m). A total of 6 images are acquired for each activation, with a 15-second wait time between the images to allow for traffic movement across the intersection. If more than one transit vehicle approach and activate the camera at the same time, a single acquisition stream is initiated and separate images will be labeled by the trip ID and direction of each transit vehicle that crossed the area during the given timeframe. The outputs of the data acquisition process are a database containing the trip IDs, directions, and the timestamp of the transit vehicles' approach to the site of study, and an image dataset with 6 images associated with each trip ID. The data acquisition process for this pilot study is summarized by the following:
\begin{itemize}
    \item Three separate rounds of data acquisition (to account for the seasonal variations in daylight and roadside greenery).
    \begin{itemize}
        \item Feb 23\textsuperscript{rd} - March 4\textsuperscript{th}, 2022 
        \item March 28\textsuperscript{th} - April 6\textsuperscript{th}, 2022
        \item May 9\textsuperscript{th} - May 16\textsuperscript{th}, 2022
    \end{itemize}
    \item Data was collected for each day between 6 AM - 9 PM.
    \item Transit trip data is only recorded if the image acquisition process is successful. Livestream buffering and connection issues can result in failed acquisitions.
    \item A total of 2,992 MBTA Route 1 bus trips were recorded.
    \item A total of 17,905 images were acquired and associated with these trips.
\end{itemize}

\begin{table}[!ht]
  \renewcommand{\arraystretch}{0.60}
  \centering
  \caption{Data Sources and Attributes Used}
    \begin{tabular}{lccc}
    \toprule
    \textbf{Source} & \textbf{Attribute} & \textbf{Type} & \textbf{Description} \\
    \midrule
    \multirow{3}[6]{*}{\textbf{GTFS-RT}} & id    & Integer & Unique trip identifier. \\
\cmidrule{2-4}          & dir   & Binary & Trip Direction (0 = outbound, 1 = inbound). \\
\cmidrule{2-4}          & timestamp & Integer & Unix timestamp during image acquisition. \\
    \midrule
    \multirow{2}[4]{*}{\textbf{AVL}} & ts    & Float & Total travel time across the segment (sec). \\
\cmidrule{2-4}          & dwell & Float & Stop dwell time for a given trip (sec). \\
    \midrule
    \textbf{Camera} & site image & 1280 x 720 x 3 Array & Image of the observed scene. \\
    \bottomrule
    \end{tabular}%
  \label{tab:sources}%
\end{table}%

\vspace{-10pt}
\subsubsection{Generating Travel Time Bands Based on Effective Travel Time Percentiles}
The acquired trip-image dataset requires pre-processing to make it suitable for the ViT image classification task. First, the overall travel time of transit vehicles is acquired from the MBTA's AVL database. The AVL database also keeps a record of transit vehicles' stop events, including the associated dwell time with a stop event (if any). We associate the trip ID for each transit vehicle in our dataset with the stop events at either of the two stops in our site of study (in the inbound and outbound direction) and define the effective travel time as the time spent by the vehicles on the 1-km segment (which includes camera monitored area in addition to the 500m activation buffer in either direction), minus any dwell time associated with that trip at its respective stop on the site.

\begin{figure}[!ht]
\centering
\includegraphics[width= 0.58\textwidth]{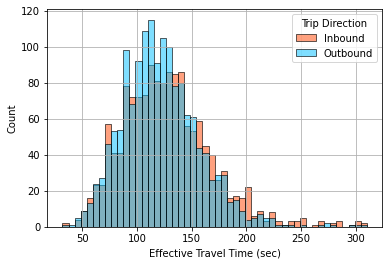}
\caption{\label{fig:travel} Observed effective transit travel time for training dataset.}
\end{figure}

\begin{figure}[!ht]
\begin{subfigure}{.49\textwidth}
  \centering
  \includegraphics[width= \textwidth, height = 2.2in]{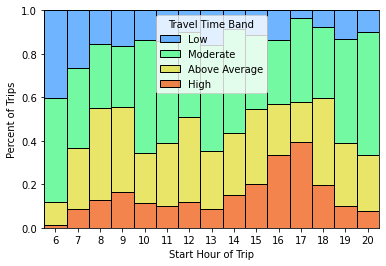}
  \caption{\label{fig:inbound_dist} Inbound.}
\end{subfigure}
\begin{subfigure}{.49\textwidth}
  \centering
  \includegraphics[width= \textwidth, height = 2.2in]{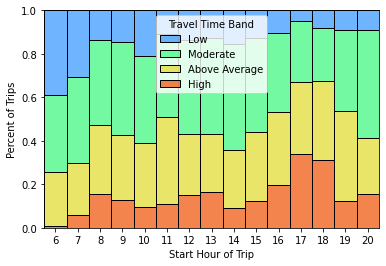}
  \caption{\label{fig:outbound_dist} Outbound.}
\end{subfigure}
\caption{\label{fig:travel_time_distribution} Travel time band distribution for acquired image dataset.}
\end{figure}

We use this effective travel time across the segment as a ground truth label, with the assumption that after accounting for stop events (if any) a transit vehicle's travel time across a segment is representative of the overall traffic. This assumption, however, doesn't take into account the impact of signal control due to the unavailability of signal time data. The signal, however, was observed to operate a 180-second cycle length, with an 85-15 split (25 seconds green for the side street) which considerably favors the street on which Route 1 transit vehicles run. The observed effective travel time for the dataset is illustrated in Figure \ref{fig:travel}. Figure \ref{fig:travel_time_distribution} shows the distribution of these travel time bands per trip direction and hour of the day for the acquired image sequence dataset. It is worth noting that the dataset contains more outbound (1,536) than inbound (1,456) trips.

The overall average effective travel time observed was 124 seconds, with a minimum of 35 seconds and a maximum of 310 seconds. The descriptive statistics are shown in Table \ref{tab:stats}. To normalize the labels for the training dataset, we utilize a four-class labeling approach to discretize observed effective travel time percentile and label them as follows:
\begin{itemize}
    \item Effective Travel Time $\leq$ 10\% $\rightarrow$ Low.
    \item 10\% $<$ Effective Travel Time $\leq$ 50\% $\rightarrow$ Moderate.
    \item 50\% $<$ Effective Travel Time $<$ 90\% $\rightarrow$ Above Average.
    \item Effective Travel Time $\geq$ 90\% $\rightarrow$ High.
\end{itemize}

\begin{table}[!ht]
  \renewcommand{\arraystretch}{0.71}
  \centering
  \caption{Descriptive Statistics for Effective Travel Time}
    \begin{tabular}{lccc}
    \toprule
    \multirow{2}[4]{*}{\textbf{Statistic}} & \multicolumn{3}{c}{\textbf{Trip Direction}} \\
\cmidrule{2-4}          & \textbf{Overall} & \textbf{Inbound} & \textbf{Outbound} \\
    \midrule
    \textbf{$\mu$} & 124   & 127   & 121 \\
    \midrule
    \textbf{$\sigma$} & 38    & 41    & 35 \\
    \midrule
    \textbf{10\%} & 79    & 79    & 79 \\
    \midrule
    \textbf{50\%} & 121   & 124   & 118 \\
    \midrule
    \textbf{90\%} & 160   & 166   & 156 \\
    \midrule
    \textbf{Min} & 35    & 35    & 35 \\
    \midrule
    \textbf{Max} & 310   & 309   & 310 \\
    \midrule
    \textbf{Count} & 2,992  & 1,456  & 1,536 \\
    \bottomrule
    \end{tabular}%
  \label{tab:stats}%
\end{table}%

\subsubsection{ViT Model Training and Fine-tuning}
The ViT model utilized as a part of our framework is the one proposed by Dosovitskiy et al. \cite{dosovitskiy2020image}. The gist of the model is that it simplifies the pixel-wise attention calculation that would take place in a transformer's encoder module by splitting the image into $N$ fixed-size patches ($P$). Those fixed-size patches are linearly inserted into the transformer encoder alongside their positional embeddings, which simply tell the encoder where each patch belongs in an image. Meaning that an input image \textbf{x} $\in \mathbb{R}^{H \times W \times C}$ is reshaped into \textbf{x}\textsubscript{p} $\in \mathbb{R}^{N \times (P^{2 \ . \ } C)}$, where H, W, C are respectively the height (720 pixels), width (1280 pixels), and the number of color channels (3) in the source image. A constant latent vector ($D$) is used across all layers, and learnable positional embedding parameters (\textbf{E}$_{lin}$, \textbf{E}$_{pos}$) are utilized to extract an embedding token (\textbf{z}$_o$). Flattened embeddings pass through a series of encoders ($L$) each with a multi-head self-attention (MSA) layer, and a feed-forward Multi-Layer Perceptron (MLP), both preceded by normalization layers (LN). This process is mathematically described by the following:

\begin{align}
    \textbf{z}_o &= [\textbf{x}_{class};\ \textbf{x}_{p}^{1}\textbf{E};\ \textbf{x}_{p}^{2}\textbf{E};\ ... \ ;\ \textbf{x}_{p}^{N}\textbf{E}] + \textbf{E}_{pos}, &\textbf{E}\in\mathbb{R}^{(P^{2 \ . \ } C) \times D},\ \textbf{E}_{pos} \in \mathbb{R}^{(N + 1) \times D}\\
    \textbf{z}_{l}^{'} &= \text{MSA}(\text{LN}(\textbf{z}_{l-1})) + \textbf{z}_{l-1}, & l= 1\ ... \ L\\
    \textbf{z}_{l} &= \text{MLP}(\text{LN}(\textbf{z}_{l}^{'})) + \textbf{z}_{l}^{'}, & l= 1\ ... \ L\\
    \textbf{y} &= \text{LN}(\textbf{z}_{L}^{0})
\end{align}

\noindent Where \textbf{y} is the final image representation out of the transformer encoder, then passed into a classification MLP outputting the most likely class for the image, which in our case is the travel time band prediction based on the effective travel time percentile. We start with the base ViT model pre-trained on the ImageNet-21K dataset and open-sourced by Google Research \cite{dosovitskiy2020image}. To accommodate for the data-hungry nature of ViT, prior to re-training on our data we increase the size of the dataset by creating an image augmentation pipeline; randomly cropping, tilting, and slightly adjusting the brightness and contrast of images acquired in our dataset as shown in Table \ref{tab:augmentation}, while maintaining the original images' class label based on effective travel time rank. Every image is passed through the pipeline 6 times, and the probability of an augmentation action is the probability that it is performed on the image during an iteration in a random magnitude within action bounds. This results in a final dataset with 78,385 images.

\begin{table}[!ht]
  \renewcommand{\arraystretch}{1.20}
  \centering
  \caption{Image Augmentation Pipeline Parameters}
    \begin{tabular}{lccc}
    \toprule
    \multirow{2}[4]{*}{\textbf{Action}} & \multicolumn{2}{c}{\textbf{Augmentation Bounds}} & \multirow{2}[4]{*}{\textbf{Probability}} \\
\cmidrule{2-3}          & \textbf{Lower} & \textbf{Upper} &\\
    \midrule
    \textbf{Crop} & 560 $\times$ 560 & 1280 $\times$ 720 & 0.33 \\
    \midrule
    \textbf{Rotate} & -30\textsuperscript{ o}  & +30\textsuperscript{ o}    & 0.33 \\
    \midrule
    \textbf{Brightness} & -20\% & +20\%  & 0.50 \\
    \midrule
    \textbf{Contrast} & -20\% & +20\%  & 0.50 \\
    \bottomrule
    \end{tabular}%
  \label{tab:augmentation}%
\end{table}%

We initially train multiple instances of the ViT model while performing a grid search to optimize the model's hyper-parameters to minimize the overall F-1 Score. The bounds and optimal values of this fine-tuning process are shown in Table \ref{tab:vit_param}. After finding optimal hyper-parameters, the vision transformer was trained using a 5-fold cross-validation, with an 80-20 stratified training-testing split for each fold. We further evaluate the model in terms of precision, recall, and accuracy.

\begin{table}[!ht]
  \renewcommand{\arraystretch}{1.20}
  \centering
  \caption{ViT Parameter Fine-Tuning}
    \begin{tabular}{lccc}
    \toprule
    \multirow{2}[4]{*}{\textbf{Parameter}} & \multicolumn{2}{c}{\textbf{Parameter Limits}} & \multicolumn{1}{c}{\multirow{2}[4]{*}{\textbf{Optimal}}} \\
\cmidrule{2-3}          & \textbf{Lower} & \textbf{Upper} &  \\
    \midrule
    \textbf{Hidden Layers} & 2     & 12   & 12 \\
    \midrule
    \textbf{Attention Heads} & 2     & 12    & 12 \\
    \midrule
    \textbf{Batch Size} & 8     & 256   & 32 \\
    \midrule
    \textbf{Learning Rate} & 1$e^{-6}$ & 1$e^{-2}$ & 2$e^{-5}$ \\
    \midrule
    \textbf{Dropout Probability} & 0 & 0.25     & 0.10 \\
    \bottomrule
    \end{tabular}%
  \label{tab:vit_param}%
\end{table}%

\section{Results and Discussion}
\subsection{ViT Performance on Effective Travel Time Band Prediction}
\begin{figure}[!ht]
\begin{subfigure}{.49\textwidth}
  \centering
  \includegraphics[width= \textwidth, height = 2.4in]{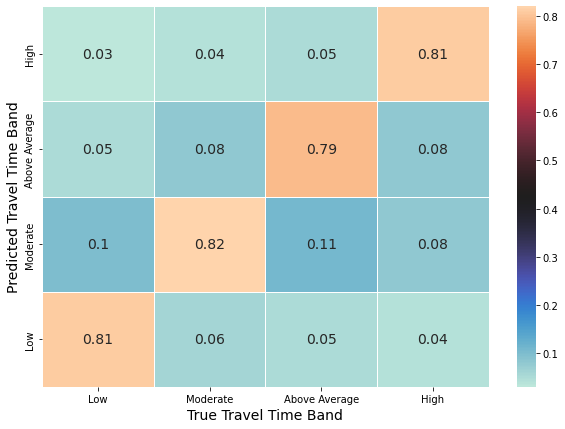}
  \caption{\label{fig:cm_in_p} Inbound.}
\end{subfigure}
\begin{subfigure}{.49\textwidth}
  \centering
  \includegraphics[width= \textwidth, height = 2.4in]{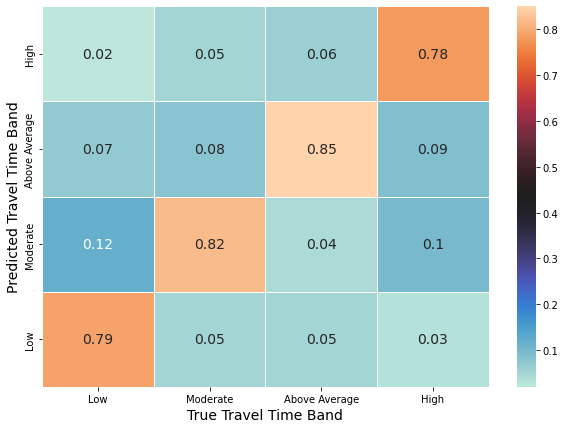}
  \caption{\label{fig:cm_out_p} Outbound.}
\end{subfigure}
\caption{\label{fig:confusion_p} Normalized confusion matrices for test subsets.}
\end{figure}

After training on 80\% of the images in the k-fold augmented dataset, the averaged 5-fold performance of the optimized ViT model on the test sets for either direction is shown in the confusion matrices in Figure \ref{fig:confusion} below, illustrating the true and predicted travel time band for test images. Figure \ref{fig:confusion_p} illustrates the results normalized to the total number of true travel time band images in the test set (sum over columns = 1, diagonal cells represent accuracy for each travel time band). 

\begin{figure}[!ht]
\begin{subfigure}{.49\textwidth}
  \centering
  \includegraphics[width= \textwidth, height = 2.4in]{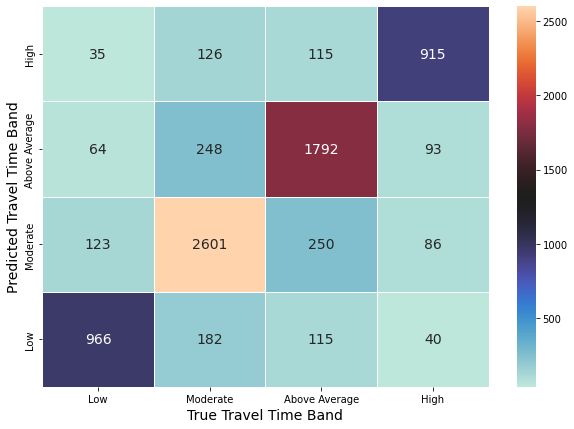}
  \caption{\label{fig:cm_in} Inbound.}
\end{subfigure}
\begin{subfigure}{.49\textwidth}
  \centering
  \includegraphics[width= \textwidth, height = 2.4in]{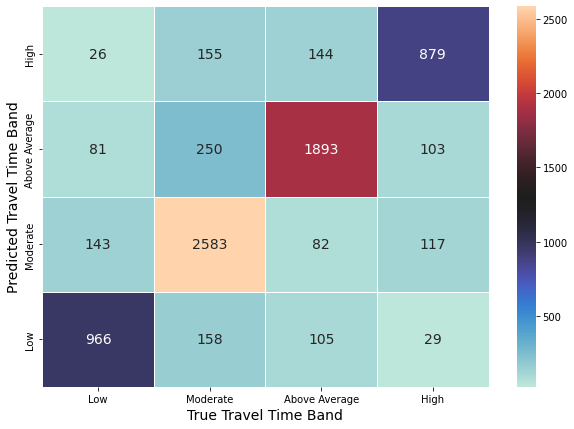}
  \caption{\label{fig:cm_out} Outbound.}
\end{subfigure}
\caption{\label{fig:confusion} Confusion matrices for test subsets.}
\end{figure}

The results indicate that the model is able to differentiate the discrete expected travel time ranges based on the observed effective transit travel time used for labeling the dataset. While a higher number of misclassifications happen between the intermediate states (moderate and above-average conditions), the misclassification rate between the extreme conditions (low and high) is very low. This result is of utmost importance, indicating that when using image data in complementing the travel time prediction process, the misclassifications will rarely lead to an extreme-end under or overestimation. Detailed results on the ViT model's performance are provided in Table \ref{tab:test}, showing the 5-folds average for each metric $\scriptstyle \pm$ the range of variation for each metric across the 5-folds.



\begin{table}[!ht]
  \renewcommand{\arraystretch}{0.9}
  \centering
  \caption{5-Fold Cross-Validation Test Sets Performance Metrics}
    \begin{tabular}{lcccccc}
    \toprule
    \multicolumn{1}{l}{\multirow{2}[4]{*}{\textbf{Class}}} & \multicolumn{3}{c}{\textbf{Inbound}} & \multicolumn{3}{c}{\textbf{Outbound}} \\
\cmidrule{2-7}          & \textbf{Prc $\scriptstyle (\pm)$} & \textbf{Recall $\scriptstyle (\pm)$} & \textbf{F-1 $\scriptstyle (\pm)$} & \textbf{Prc $\scriptstyle (\pm)$} & \textbf{Recall $\scriptstyle (\pm)$} & \textbf{F-1 $\scriptstyle (\pm)$} \\
    \midrule
    \textbf{Low} & 0.74 $\scriptstyle(0.05)$  & 0.81 $\scriptstyle(0.02)$  & 0.78 $\scriptstyle(0.01)$ & 0.76 $\scriptstyle(0.02)$  & 0.80 $\scriptstyle(0.02)$  & 0.79 $\scriptstyle(0.03)$ \\
    \midrule
    \textbf{Normal} & 0.84 $\scriptstyle(0.01)$  & 0.83 $\scriptstyle(0.02)$  & 0.83 $\scriptstyle(0.01)$ & 0.82 $\scriptstyle(0.01)$  & 0.82 $\scriptstyle(0.01)$  & 0.82 $\scriptstyle(0.01)$\\
    \midrule
    \textbf{Above Average} & 0.81 $\scriptstyle(0.01)$  & 0.78 $\scriptstyle(0.01)$ & 0.80 $\scriptstyle(0.00)$ & 0.81 $\scriptstyle(0.01)$  & 0.79 $\scriptstyle(0.02)$  & 0.79 $\scriptstyle(0.01)$ \\
    \midrule
    \textbf{High} & 0.77 $\scriptstyle(0.02)$  & 0.81 $\scriptstyle(0.02)$ & 0.79 $\scriptstyle(0.01)$ & 0.74 $\scriptstyle(0.02)$  & 0.79 $\scriptstyle(0.03)$  & 0.75 $\scriptstyle(0.03)$ \\
    \toprule
    \textbf{Accuracy} & \multicolumn{3}{c}{\textbf{0.81 $\scriptstyle(0.01)$}} & \multicolumn{3}{c}{\textbf{0.80 $\scriptstyle(0.01)$}} \\
    \bottomrule
    \end{tabular}%
  \label{tab:test}%
\end{table}%

\begin{figure}[!ht]
\begin{subfigure}{.49\textwidth}
  \centering
  \includegraphics[width= \textwidth, height = 2.4in]{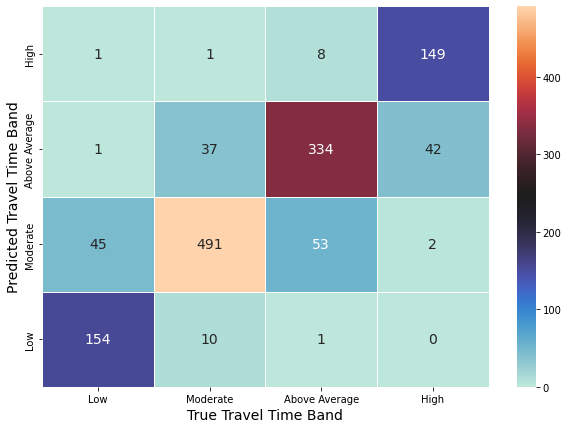}
  \caption{\label{fig:cm_in_av} Inbound.}
\end{subfigure}
\begin{subfigure}{.49\textwidth}
  \centering
  \includegraphics[width= \textwidth, height = 2.4in]{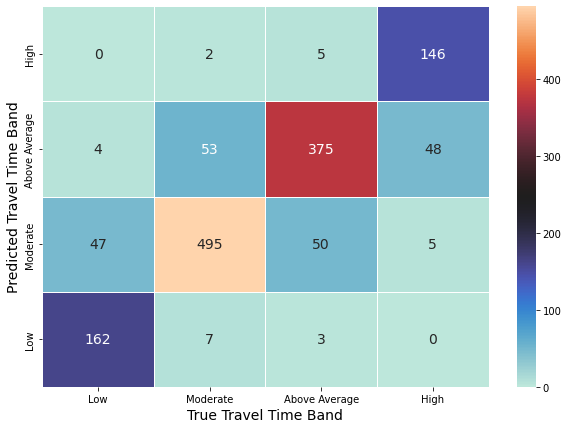}
  \caption{\label{fig:cm_out_av} Outbound.}
\end{subfigure}
\caption{\label{fig:av_confusion} Confusion matrices using averaged image sequence score.}
\end{figure}

\begin{figure}[!ht]
\begin{subfigure}{.49\textwidth}
  \centering
  \includegraphics[width= \textwidth, height = 2.4in]{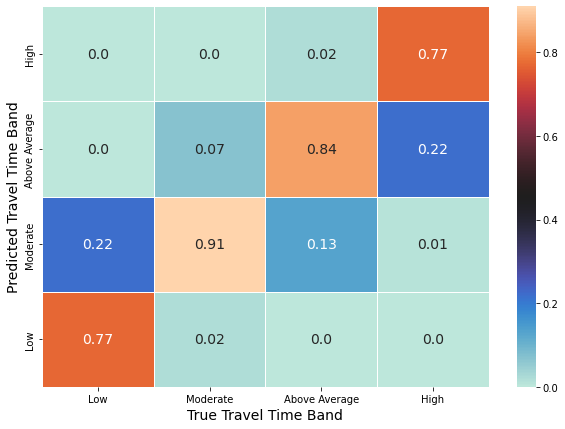}
  \caption{\label{fig:cm_in_av_p} Inbound.}
\end{subfigure}
\begin{subfigure}{.49\textwidth}
  \centering
  \includegraphics[width= \textwidth, height = 2.4in]{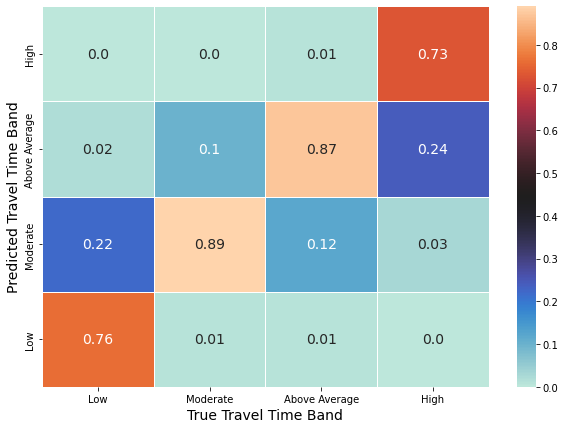}
  \caption{\label{fig:cm_out_av_p} Outbound.}
\end{subfigure}
\caption{\label{fig:av_confusion_p} Normalized confusion matrices using averaged image sequence score.}
\end{figure}

\noindent Given that the ViT training and inference are based on images that provide a constrained view of the area of study, variation in the number of vehicles and other factors between images taken from the same sequence can occur (6 images at 15-second intervals, as detailed in the Methodology). To account for this variation, we run inference on image sequences for a given trip ID instead of single images. Images from the test set of the best-performing fold were grouped by their trip ID before making travel time band predictions based on the average of all predictions for the image sequence. The results are illustrated in Figures \ref{fig:av_confusion} and \ref{fig:av_confusion_p}. Classification metrics are detailed in Table \ref{tab:averaged}, where support is the number of actual occurrences of the class in the test subset.

\begin{table}[!ht]
  \renewcommand{\arraystretch}{0.9}
  \centering
  \caption{Performance Metrics Using Averaged Image Sequence Score}
    \begin{tabular}{lcccccccc}
    \toprule
    \multicolumn{1}{c}{\multirow{2}[4]{*}{\textbf{Class}}} & \multicolumn{4}{c}{\textbf{Inbound}} & \multicolumn{4}{c}{\textbf{Outbound}} \\
\cmidrule{2-9}          & \textbf{Prc} & \textbf{Recall} & \textbf{F-1} & \textbf{Support} & \textbf{Prc} & \textbf{Recall} & \textbf{F-1} & \textbf{Support} \\
    \midrule
    \textbf{Low} & 0.93  & 0.77  & 0.84  & 201   & 0.94  & 0.76  & 0.84  & 213 \\
    \midrule
    \textbf{Moderate} & 0.83  & 0.91  & 0.87  & 539   & 0.83  & 0.89  & 0.86  & 557 \\
    \midrule
    \textbf{Above Average} & 0.81  & 0.84  & 0.82  & 396   & 0.78  & 0.87  & 0.82  & 433 \\
    \midrule
    \textbf{High} & 0.94  & 0.77  & 0.85  & 193   & 0.95  & 0.73  & 0.83  & 199 \\
    \toprule
    \textbf{Accuracy} & \multicolumn{4}{c}{\textbf{0.85}} & \multicolumn{4}{c}{\textbf{0.84}} \\
    \bottomrule
    \end{tabular}%
  \label{tab:averaged}%
\end{table}%

\newpage
The results in Table \ref{tab:averaged} demonstrate significant improvements in model performance when using image sequences, particularly in nearly eliminating all misclassifications between non-consecutive travel time ranges. While the accuracy (the number of travel time band labels that were correctly classified divided by the total occurrences of the class label in the training dataset) slightly drops for the under-represented classes (low and high travel time bands), the precision for those classes increases significantly. Given that precision is an indicator of the quality of the prediction (the number of true positives divided by the total number of class predictions made by the model) a precision of over 93\% for those extreme classes is of utmost importance when those labels are to be used to enhance travel time predictions. The precision, recall, F-1 score, and accuracy all increase for the moderate and above-average travel time band labels.

\begin{figure}[!ht]
\centering
\includegraphics[width= 0.55\textwidth]{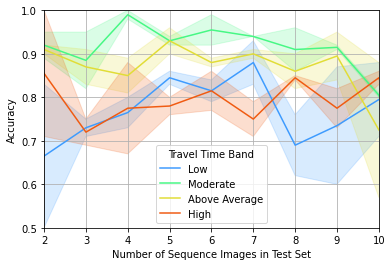}
\caption{\label{fig:image_count} Accuracy versus image count in a test sequence.}
\end{figure}

Of the 2,992 trip-image sequences in the dataset acquired for this study, 2,731 transit trip IDs had two or more images in the test set of the best-performing training fold. The effect of the number of images in a sequence is illustrated in Figure \ref{fig:image_count}. The line plots show the average classification accuracy for each travel time band, while the envelopes illustrate the 95\textsuperscript{th} confidence interval for a given number of images in sequence. While accuracy and confidence increase as the number of images used for prediction increases, particularly for under-represented classes (the low and high travel time bands), a decline is observed after the number of images in a sequence exceeds 6. This is attributable to the original number of images for each trip-image sequence acquisition being 6 images, hence exceeding that number indicates the presence of augmented images in a test sequence that could make it more challenging to classify.

\begin{figure}[!ht]
\begin{subfigure}{.49\textwidth}
  \centering
  \includegraphics[width= 0.95\textwidth, height = 2.4in]{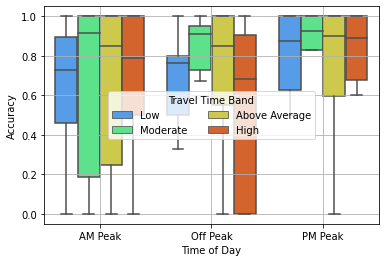}
  \caption{\label{fig:pred_dist_inbound} Inbound.}
\end{subfigure}
\begin{subfigure}{.49\textwidth}
  \centering
  \includegraphics[width= 0.95\textwidth, height = 2.4in]{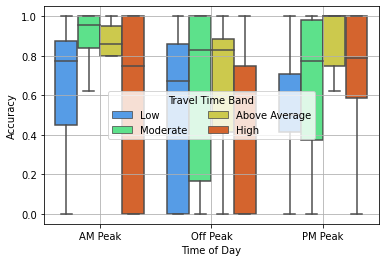}
  \caption{\label{fig:pred_dist_outbound} Outbound.}
\end{subfigure}
\caption{\label{fig:prediction_distribution} Variation in classification accuracy by direction and time of day.}
\end{figure}

Figure \ref{fig:prediction_distribution} illustrates the variation in travel time band prediction accuracy for both directions across different times of the day. Consistent with previously discussed results, the model's performance for the inbound direction outperforms the outbound. The model was found to learn and make better predictions during the AM and PM peak hours for both directions compared to the off-peak hours. This may be attributed to the apparent variation in traffic volumes observed during peak hours which is a good indicator of the expected travel time across the monitored segment during these times. Higher travel times can occur during off-peak hours without the presence of high traffic due to traffic control, driving behavior, or vehicles impeding access to bus stops. Such factors can not be identified from the image data used in this study.

\begin{figure}[!ht]
\begin{subfigure}{.5\textwidth}
  \centering
  \includegraphics[width= \textwidth]{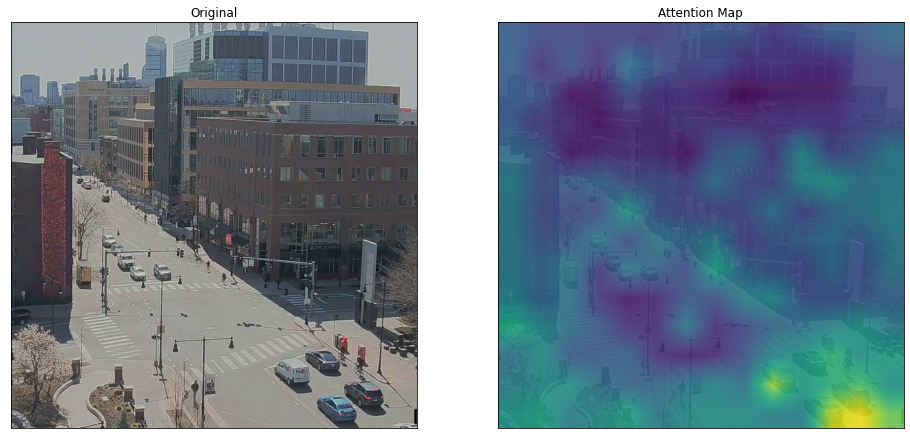}
  \caption{Normal.}
\end{subfigure}
\begin{subfigure}{.5\textwidth}
  \centering
  \includegraphics[width= \textwidth]{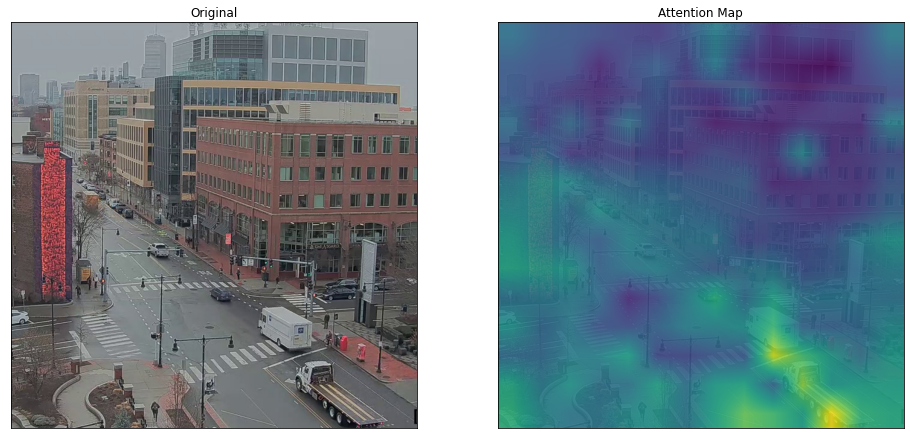}
  \caption{Rain.}
\end{subfigure}\\
\begin{subfigure}{.5\textwidth}
  \centering
  \includegraphics[width= \textwidth]{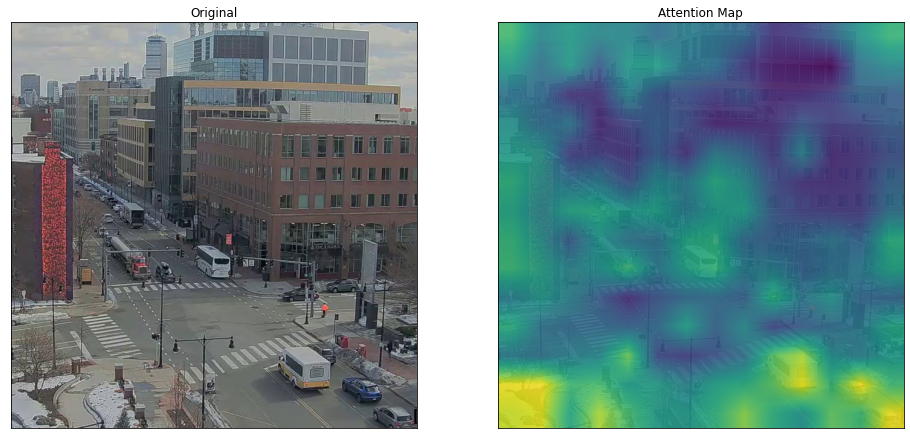}
  \caption{\label{fig:inbound_snow}Snow.}
\end{subfigure}
\begin{subfigure}{.5\textwidth}
  \centering
  \includegraphics[width= \textwidth]{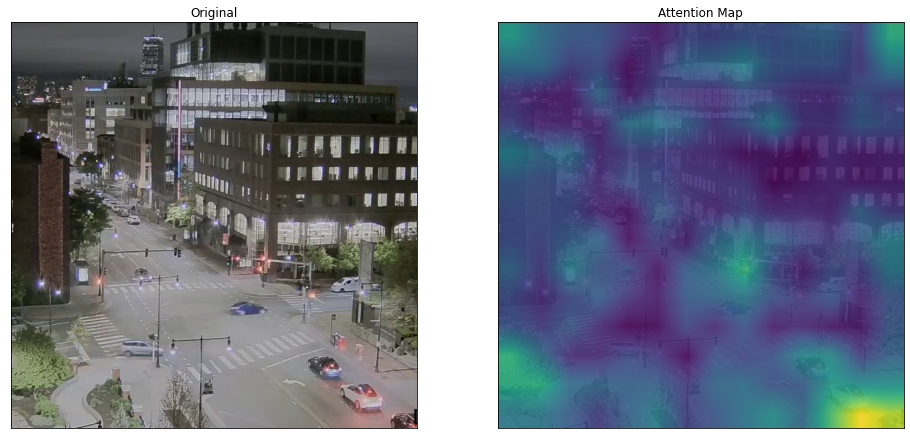}
  \caption{\label{fig:inbound_night}Night.}
\end{subfigure}
\caption{\label{fig:inbound_attention} Inbound direction attention maps across different conditions.}
\end{figure}

\begin{figure}[!ht]
\begin{subfigure}{.499\textwidth}
  \centering
  \includegraphics[width= \textwidth]{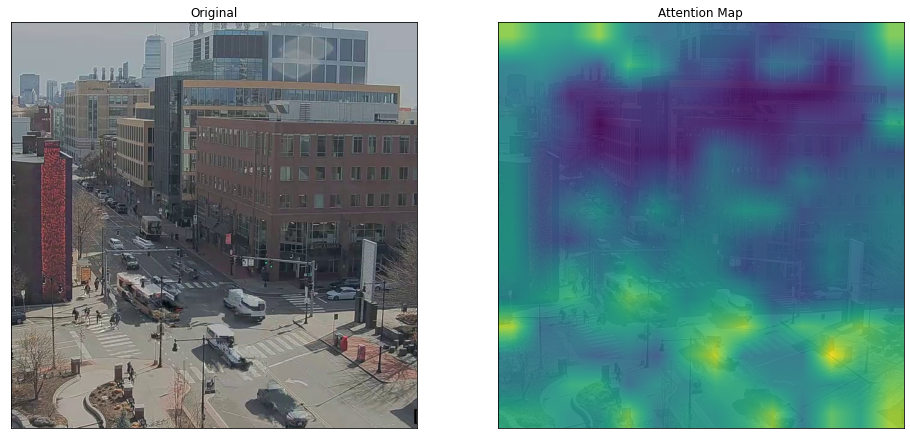}
  \caption{Normal.}
\end{subfigure}
\begin{subfigure}{.5\textwidth}
  \centering
  \includegraphics[width= \textwidth]{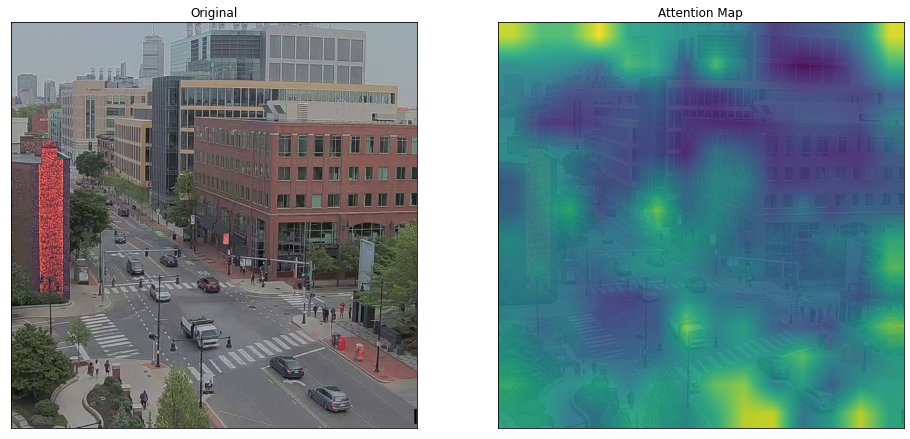}
  \caption{Rain.}
\end{subfigure}\\
\begin{subfigure}{.5\textwidth}
  \centering
  \includegraphics[width= \textwidth]{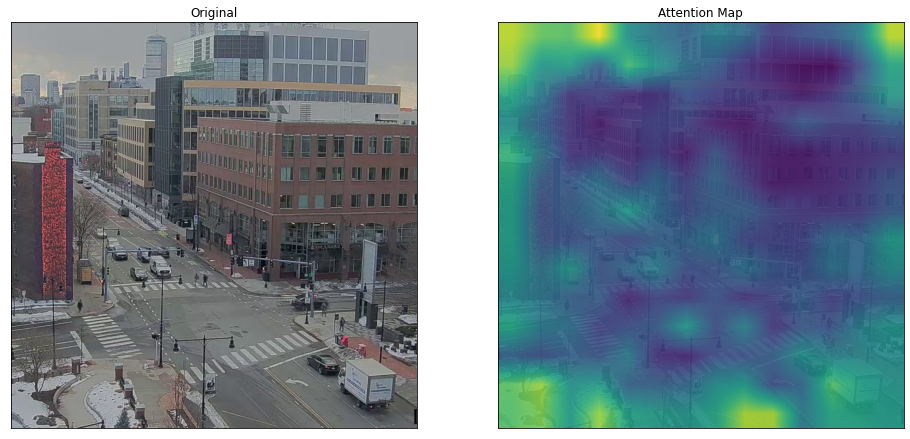}
  \caption{Snow.}
\end{subfigure}
\begin{subfigure}{.5\textwidth}
  \centering
  \includegraphics[width= \textwidth]{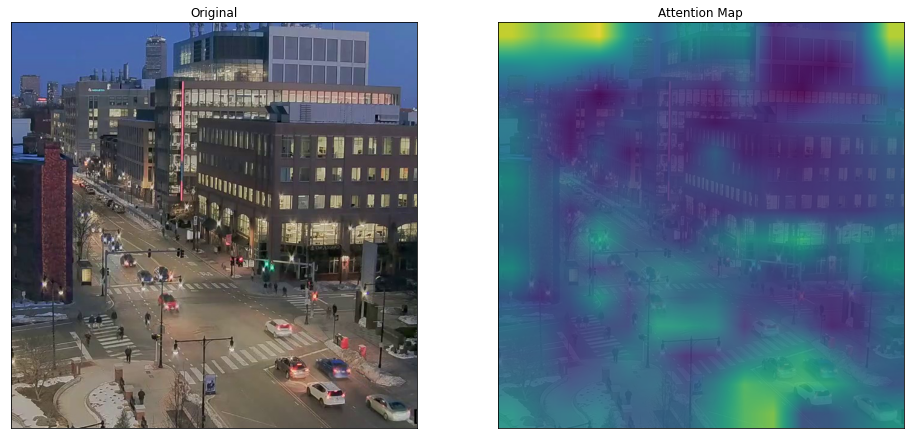}
  \caption{Night.}
\end{subfigure}
\caption{\label{fig:outbound_attention} Outbound direction attention maps across different conditions.}
\end{figure}

\newpage
Next, we assess the interpretability of the results obtained by the ViT. This is accomplished by mapping the averaged attention scores (values between 0 to 1) of the model's 12 attention heads. Attention illustrates the parts of the image from which the model learns to gain the most information in making the predictions for travel time bands. Higher scores are illustrated by brighter pixels. Figures \ref{fig:inbound_attention} and \ref{fig:outbound_attention} illustrate the attention maps for different scenarios for the inbound and outbound directions.

\newpage
We observe the ViT model learning to make predictions in a very logical and sensible manner, where a lot of attention is directed to the pixels containing information about vehicles traveling in the direction of interest. For the inbound direction illustrated in Figure \ref{fig:inbound_attention}, we see the majority of attention being directed to the right side of the image, mostly to the vehicle in the queue in the inbound direction (when present), but also towards the downstream. The ViT learns to look in the opposing direction with lower attention while learning to ignore the buildings that provide little to no information (although attention to buildings' indoor lights was observed, the model might use it as an indicator for the time of day). Other sensible actions illustrated by the ViT include paying attention to the snow when there is a snowfall, which impacts the travel speeds across the segment. In Figure \ref{fig:inbound_night}, it can be seen that at night the transformer learns to include information from the inbound direction traffic light which becomes visible from the camera's vantage point in the evening.

We see the ViT model exhibiting similar behavior in the outbound direction in Figure \ref{fig:outbound_attention}. The camera view for the outbound direction, however, is a lot more limited compared to that of the inbound, as can be seen with the building on the left of the image blocking the upstream, and a very limited view of the downstream. This explains the better predictive performance for the model in the inbound direction previously presented in the evaluation tables and confusion matrices. Asides from learning to look at vehicles and their locations, the model was observed to pay more attention to the sky and buildings' lights during the night. This behavior seems to be an attempt to compensate for the lack of information induced by the restricted camera view for the outbound direction by making some form of time-of-day inference.

\subsection{Implications for Travel Time Prediction}
\begin{figure}[!ht]
\begin{subfigure}{.49\textwidth}
  \centering
  \includegraphics[width= 0.92\textwidth]{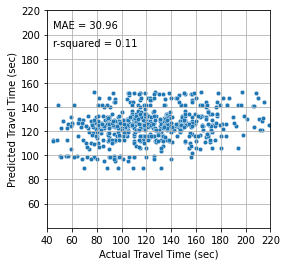}
  \caption{\label{fig:act_pred} AVL-based linear regression.}
\end{subfigure}
\begin{subfigure}{.49\textwidth}
  \centering
  \includegraphics[width= 0.92\textwidth]{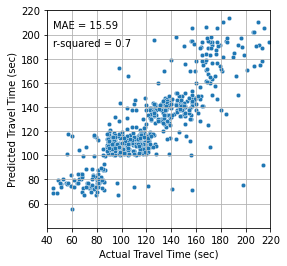}
  \caption{\label{fig:act_pred_tranvit} With image-based band prediction.}
\end{subfigure}
\caption{\label{fig:travel_time_prediction} Exploring a linear regression fit for predicting transit travel time through a segment.}
\end{figure}

We conclude this assessment with a brief proof-of-concept looking into what this use of real-time, computer-vision-based travel time class predictions could mean to the broader task of travel and arrival time estimation. We fit a linear regression model to predict the effective segment travel time of the 2,731 vehicles in the image sequence test set based on the recorded information for these vehicles which was obtained from GTFS real-time component. The linear regression model (Ordinary Least Squares, OLS) makes baseline travel time predictions based on the time of day, the direction of travel, and the occupancy of the transit vehicle as it approaches the segment. Occupancy is a continuous variable showing the percentage of passengers estimated from on-bus automatic passenger counters to total seating capacity, ranges from 0 - 150\%, with $\mu = 34.6$ and $\sigma = 26.67$. Hours of the day were encoded as binary variables. We then run the same model, with the addition of a predicted travel time band label (OLS+) obtained from the inference of the images associated with the transit vehicle's approach to the segment. The predicted travel time bands are denoted by the TTB variables, with the moderate band (TTB\_Mod) as the baseline. The results of actual versus predicted travel times are illustrated in Figure \ref{fig:travel_time_prediction}.

Table \ref{tab:ols} shows the coefficient estimates for the different models. Hours of the day that were not found significant for any model were dropped from the table for brevity. The model-agnostic addition of the travel time band labels predicted from images works to create bounds that enhance the continuous travel time estimation. This demonstratory evaluation shows significant improvement achieved by the linear regression model, both in terms of the predictions' mean absolute error and r-squared. As the data and image sequence acquisition for this study was initiated when a transit vehicle is within 500 meters of the camera, we utilized state predictions from the previous transit vehicle traveling in the same direction and obtained comparable results, indicating that the traffic state does not change drastically between consecutive vehicles (10-minute scheduled headway for MBTA Route 1). Future work will investigate the extent to which this look-ahead horizon can provide reliable predictions, and assess the extent of performance improvements to more sophisticated state-of-the-practice models.

\begin{table}[!ht]
  \centering
  \renewcommand{\arraystretch}{0.60}
  \caption{Linear Regression Model Estimates}
    \begin{tabular}{lccccc}
    \toprule
    \multicolumn{1}{c}{\multirow{2}[4]{*}{\textbf{Variable}}} & \multirow{2}[4]{*}{\textbf{Type}} & \multicolumn{2}{c}{\textbf{Inbound}} & \multicolumn{2}{c}{\textbf{Outbound}} \\
\cmidrule{3-6}          &       & \textbf{OLS} & \textbf{OLS+} & \textbf{OLS} & \textbf{OLS+} \\
    \toprule
    \textbf{Occupancy} & \textbf{Continous} & 0.23$^{**}$ & 0.04  & 0.02  & 0.03 \\
    \midrule
    \textbf{Hour\_6} & \textbf{Binary} & -25.11$^{**}$ & 0.16  & -15.63$^{**}$ & 1.93 \\
    \midrule
    \textbf{Hour\_7} & \textbf{Binary} & -12.92$^{**}$ & 2.80  & -15.79$^{**}$ & 2.84 \\
    \midrule
    \textbf{Hour\_9} & \textbf{Binary} & 8.53* & 4.91$^{*}$  & 11.54$^{**}$ & 5.76$^{**}$ \\
    \midrule
    \textbf{Hour\_12} & \textbf{Binary} & 6.12  & 0.29 & 11.10$^{**}$ & 7.15$^{**}$ \\
    \midrule
    \textbf{Hour\_13} & \textbf{Binary} & 1.49  & 2.50  & 2.49  & 5.40$^{*}$ \\
    \midrule
    \textbf{Hour\_14} & \textbf{Binary} & -2.47  & 3.77  & 7.42$^{**}$ & 7.91$^{**}$ \\
    \midrule
    \textbf{Hour\_16} & \textbf{Binary} & 21.79$^{**}$ & 11.69$^{**}$ & 7.04$^{**}$ & 3.69 \\
    \midrule
    \textbf{Hour\_17} & \textbf{Binary} & 24.85$^{**}$ & 6.46$^{*}$ & 22.47$^{**}$ & 8.03$^{**}$ \\
    \midrule
    \textbf{Hour\_18} & \textbf{Binary} & 25.10$^{**}$ & 11.57$^{**}$ & 37.19$^{**}$ & 15.88$^{**}$ \\
    \midrule
    \textbf{Hour\_19} & \textbf{Binary} & 3.11  & 4.57 & 18.18$^{**}$ & 9.64$^{**}$ \\
    \midrule
    \textbf{Hour\_20} & \textbf{Binary} & 6.14  & 10.16$^{**}$ & 11.82$^{**}$ & 6.05$^{**}$ \\
    \midrule
    \textbf{Hour\_21} & \textbf{Binary} & -8.35 & 1.24  & 9.87$^{**}$ & 8.71$^{**}$ \\
    \midrule
    \textbf{TTB\_Low} & \textbf{Binary} & -     & -33.87$^{**}$ & -     & -30.68$^{**}$ \\
    \midrule
    \textbf{TTB\_Aav} & \textbf{Binary} & -     & 36.73$^{**}$ & -     & 29.80$^{**}$ \\
    \midrule
    \textbf{TTB\_High} & \textbf{Binary} & -     & 83.32$^{**}$ & -     & 70.81$^{**}$ \\
    \midrule
    \multicolumn{2}{c}{\textit{\textbf{Intercept}}} & \textit{116.55} & \textit{104.63} & \textit{112.58} & \textit{99.87} \\
    \midrule
    \multicolumn{2}{c}{\textit{\textbf{R-Square}}}& \textit{0.111} & \textit{0.693} & \textit{0.128} & \textit{0.680} \\
    \bottomrule
    \end{tabular}%
  \label{tab:ols}%
\end{table}%

\noindent {\small Coefficients denoted with "$^{**}$" are significant at $p < 0.05$ level; "$^{*}$" are significant at $p < 0.10$}.

\section{Conclusions}
This study provided a detailed implementation and assessment for TranViT, an integrated framework utilizing a combination of traditional transit data sources with roadside computer vision to improve real-time transit travel time prediction. An exploratory assessment of our proposed framework was conducted for a segment of Massachusetts Avenue in Cambridge, MA, USA, with the results providing evidence for the potency of this framework. First, we introduce a workflow for automated roadside image data acquisition and labeling utilizing traditional transit data sources. Second, we train and thoroughly evaluate the ViT component of the framework which was able to successfully learn image features and contents that best help it deduce the expected travel time range across the segment of interest, with a validation accuracy ranging between 80\%-85\%, and a precision of up to 95\%. Finally, we demonstrate how this prediction of the travel time range can subsequently be utilized to improve continuous travel time prediction. We believe this end-to-end, scalable, automated, and highly efficient approach for integrating transit data sources and roadside imagery to extract real-time information addresses a fundamental gap impeding the broader utilization of computer vision in transit operations, which can have major implications for improving arrival time predictions and passenger real-time information.

Unlike existing studies utilizing computer vision for transportation applications, our framework doesn't require having certain data in place. On the contrary, the major contribution of this work is creating a generalized workflow for acquiring and labeling the data for the computer vision tasks, which could be extended to other use cases (e.g. focusing on areas of images containing bus stops for anticipating the dwell time). Utilizing the GTFS real-time component to initiate transit data and corresponding image sequence acquisitions results in an extremely computationally-efficient workflow, running on a single CPU and with no more than 128 MBs of RAM. Based on our observations from this study, we expect a more demanding data acquisition and training task for models to be deployed in areas where the scenery changes noticeably between seasons (like our case in Cambridge, MA). The models we trained in between rounds of data acquisition did not perform well without retraining. The subsequent model re-training, however, is faster. This is attributed to the gigantic number of parameters (86 Million in the base model) that a ViT employs, which requires substantial training to ensure generalizability. The output travel time range labels (based on the percentiles used to create these ranks) can complement any existing travel time prediction algorithm by adding a complementary real-time attribute describing the currently observed state of traffic at an area of interest, supplementing what models traditionally learn from historic AVL observations. We concluded this study with a proof of concept demonstrating the potential impact of this additional vision-based input for improving transit travel time predictions. In practice, we anticipate that a data-driven deployment of roadside cameras only in locations where high transit delays are observed would be sufficient in providing reliable travel time predictions without the need to monitor large segments of the transit network. 

One of the limitations of this study is the need to generate augmented images to satisfy the data-hungry nature of vision-transformed training. Future works will look into acquiring a larger quantity of image data, alongside integrating additional data sources that provide information on signal timing (if the work is conducted in an intersection setting) or adapting the model in ways that account for the impact of traffic control variations on segment travel times. Another potential improvement is utilizing the observed speed and/or acceleration profiles of the transit vehicles as opposed to their travel time. This could either be accomplished by logging the transit vehicles' location coordinates from GTFS-RT during the data acquisition phase, or by utilizing computer-vision-based trajectory tracking for all vehicles, which would come at the expense of higher computational load but will allow for the estimation of the true overall traffic speed and state more accurately, leading to a better quality labeling for the training data.


\section{Acknowledgements}
The authors would like to thank the MBTA for providing data access which enabled this study, and the MIT SuperCloud and Lincoln Laboratory Supercomputing Center for providing high-performance computing resources that have contributed to the research results reported in this paper.

\section{Author Contribution Statement}
The authors confirm their contribution to the paper as follows: study conception and design: A.A., J.Z.; data collection: A.A.; analysis and interpretation of results: A.A.; draft manuscript preparation: A.A., J.Z. Both authors reviewed the results and approved the final version of the manuscript. The authors do not have any conflicts of interest to declare.

\nolinenumbers
\bibliographystyle{trb}
\bibliography{references}

\end{document}